\documentclass{article}

\PassOptionsToPackage{numbers,compress}{natbib}

\usepackage[final]{neurips_2019}

\usepackage[utf8]{inputenc} 
\usepackage[T1]{fontenc}    
\usepackage{hyperref}       
\usepackage{url}            
\usepackage{booktabs}       
\usepackage{amsfonts}       
\usepackage{nicefrac}       
\usepackage{microtype}      
\usepackage{xspace}
\usepackage{adjustbox}
\usepackage{graphicx}
\usepackage{enumitem}
\usepackage[dvipsnames]{xcolor}

\newcommand{\bilstm}{\textsc{bilstm}\xspace}
\newcommand{\bilstmcrf}{\textsc{bilstm-crf}\xspace}
\newcommand{\dilatedcnn}{\textsc{dilated-cnn}\xspace}

\newcommand{\transformer}{\textsc{transformer}\xspace}

\newcommand{\transformerscrf}{\textsc{transformers-crf}\xspace}
\newcommand{\bert}{\textsc{bert}\xspace}
\newcommand{\bertcrf}{\textsc{bert-crf}\xspace}
\newcommand{\bertbilstmcrf}{\textsc{bert-bilstm-crf}\xspace}
\newcommand{\legalbert}{\textsc{legalbert}\xspace}
\newcommand{\legalbertcrf}{\textsc{legalbert-crf}\xspace}
\newcommand{\legalbertbilstmcrf}{\textsc{legalbert-bilstm-crf}\xspace}
\newcommand{\dilatedcnnscrf}{\textsc{dilated-cnns-crf}\xspace}

\newcommand{\pos}{\textsc{pos}\xspace}

\newcommand{\wordvec}{\textsc{word2vec}\xspace}
\newcommand{\wordvecword}{\textsc{w2v-word}\xspace}

\newcommand{\wordvecwordchar}{\textsc{w2v-word+char}\xspace}
\newcommand{\wordvecall}{\textsc{w2v-all}\xspace}
\newcommand{\glove}{\textsc{glove}\xspace}

\newcommand{\wordvecchar}{\textsc{w2v-all+char}\xspace}
\newcommand{\wordvecelmo}{\textsc{w2v-all+elmo}\xspace}

\newcommand{\crf}{\textsc{crf}\xspace}

\newcommand{\cnn}{\textsc{cnn}\xspace}
\newcommand{\lstm}{\textsc{lstm}\xspace}
\newcommand{\ner}{\textsc{ner}\xspace}
\newcommand{\dropout}{\textsc{dropout}\xspace}
\newcommand{\elmo}{\textsc{elmo}\xspace}
\newcommand{\conll}{\textsc{conll-2003}\xspace}

\newcommand{\fire}{
    \includegraphics[scale=0.06]{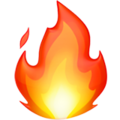}
}
\newcommand{\snowflake}{
    \includegraphics[scale=0.06]{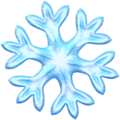}
}

\title{Neural Contract Element Extraction Revisited: \emph{Letters from Sesame Street} \\ (New updated version available only on Arxiv)$^1$}

\usepackage{authblk}
\author{\textbf{Ilias Chalkidis}}
\author{\textbf{Manos Fergadiotis}}
\author{\textbf{Prodromos Malakasiotis}}
\author{\\ \textbf{Ion Androutsopoulos}}
\affil{Institute of Informatics \& Telecommunications, NCSR ``Demokritos'', Greece \\ 
Department of Informatics, Athens University of Economics and Business, Greece}

\begin{document}

\maketitle

\begin{abstract}
  We investigate contract element extraction. We show that \lstm-based encoders perform better than dilated \cnn{s}, Transformers, and \bert in this task. We also find that domain-specific \wordvec embeddings outperform generic pre-trained \glove embeddings. Morpho-syntactic features in the form of \pos tag and token shape embeddings, as well as context-aware \elmo embeddings do not improve performance. Several of these observations contradict choices or findings of previous work on contract element extraction and generic sequence labeling tasks, indicating that contract element extraction requires careful task-specific choices. Analyzing the results of (i) plain \transformer{-based} and (ii) \bert{-based} models, we find that in the examined task, where the entities are highly context-sensitive, the lack of recurrency in \transformer{s} greatly affects their performance.\footnote{\textbf{Disclaimer:} This is an updated version that includes new results and findings that are not presented in the original version \cite{chalkidis2019} (\url{https://openreview.net/forum?id=B1x6fa95UH}).}
\end{abstract}

\section{Introduction}

Extracting information from contracts and other legal agreements is an important part of daily business worldwide. Thousands of agreements are written up every day, resulting in a huge volume of legal documents relating to several business processes, such as employment, services/vendors, loans, leases, investments. These documents contain crucial information (e.g., contract terms, pay rates, termination rights). More importantly, when negotiating or revising agreements, the parties involved need to scrutinize all the terms of the agreements as recorded in the corresponding documents.

In this work, we focus on contract element extraction, i.e., extracting information (e.g., parties, dates of interest, means of dispute, amounts) from contracts. 
As in our previous work \cite{chalkidis2017b,chalkidis2017}, the task is viewed as sequence labeling, i.e., we aim to classify each token as (possibly part of) a party name, effective date, jurisdiction, address, amount, etc.\ or `none'. However, we use a single (multi-class) classifier for all the contract elements that may reside in each contract zone (e.g., header or applicable law section), which allows the classifier of each zone to generalize across contract entity types, whereas our previous work \cite{chalkidis2017b,chalkidis2017} used a separate (binary) classifier for each contract element type per zone. Furthermore, we investigate how the following three factors affect the extractors (classifiers).

\vspace{-2mm}
\begin{itemize}[leftmargin=0in]
    \item[] {\bf Sequence encoders}: We compare several neural encoders, namely stacked \bilstm{s} \cite{Hochreiter1997}, \dilatedcnn{s} \cite{Kalchbrenner2016}, stacked \transformer{s} \cite{Vaswani2017}, and \bert \cite{devlin2019}, whereas Chalkidis et al.\ \cite{chalkidis2017b} considered only stacked \bilstm{s}. Contrary to previous studies on generic sequence labeling tasks \cite{strubell2017}, we show that \dilatedcnn{s} are not comparable to stacked \bilstm{s}. More interestingly, we show that stacked \bilstm{s} also outperform the state-of-the-art language model \bert in this task. 
    
    A combination of \bert with stacked \bilstm{s} provides comparable performance with  stacked \bilstm{s} using \wordvec embeddings, leading us to the conclusion that the lack of recurrency in \transformer{-based} models hurts the performance in the examined task, where entities are highly context-sensitive.
    
    \item[] {\bf \crf layers}: We show that the use of (linear-chain) \crf{s} \cite{Lafferty2001} on top of each encoder has a significant positive impact in all encoders,  contrary to the findings of  our previous work \cite{chalkidis2017b}, where the contribution of the \crf layer was unclear. This is most probably due to our use of multi-class classifiers, which leads to more constraints in the permissible sequences of predicted labels.
    
    \item[] {\bf Input representations}: We experiment with 200-D \glove word embeddings and domain-specific 200-D \wordvec embeddings pre-trained on approx.\ 750k contracts \cite{chalkidis2017}.\footnote{We used 200-\textsc{d} \wordvec embeddings, as in our previous work, as well as generic \glove embeddings that are available in the same dimensionality (trained on 6 billions tokens from Wikipedia and Gigaword).} We also consider additional \wordvec embeddings representing \pos tags and token shapes \cite{chalkidis2017b}, character-based word embeddings obtained by character-level \cnn{s} \cite{Ma2016},
    and context-aware \elmo embeddings \cite{peters2018}. Domain-specific \wordvec embeddings outperform generic \glove embeddings in two out of three datasets. Similar results are obtained by using \legalbert \cite{chalkidis2020}, a \bert model pre-trained on legal corpora, comparing to the original \bert model, pre-trained on generic corpora.
    Morpho-syntactic (\pos tags, token shapes), character-level, and context-aware embeddings (\elmo) 
    increase computational complexity without delivering any significant performance improvement in this task.
\end{itemize}

\section{Related Work}

\citet{Huang2015} introduced \bilstmcrf for sequence labeling (e.g., \ner, \pos tagging). \citet{Lample2016} and Ma et al.\ \cite{Ma2016} further improved the performance of \bilstmcrf by adding character-based word embeddings obtained with \bilstm and \cnn character encoders. Chiu et al.\ \cite{Chiu2016} reported mixed results by adding capitalization features and gazetteers. \citet{strubell2017} reported that \dilatedcnn{s} \cite{Kalchbrenner2016} have comparable results with \bilstm{s}, while being faster. \citet{peters2018} reported gains on several datasets, including the \conll \ner dataset, by exploiting context-aware word representations (\elmo), while \citet{devlin2019} achieved further improvements with \bert. In previous work \cite{chalkidis2017}, we introduced the task of contract element extraction, initially showing that linear window-based classifiers outperform rule-based ones. In follow up work \cite{chalkidis2017b}, we improved performance in most cases, using \lstm-based methods, with \bilstmcrf being one of the best metods. We did not, however, compare to alternative (other than \lstm-based) encoders, neither did we investigate the necessity of the morpho-syntactic features we had included in our input representations. Also, in our previous work we did not experiment with character-based word representations and context-aware word embeddings (\elmo) or pre-trained \transformer{-based} language models \cite{devlin2019, liu2019}.

\section{Task Definition and Datasets}
\label{task}

We experimented with two subsets (Header/Preamble and Applicable Law) of the publicly available data provided by Chalkidis et al.\ \cite{chalkidis2017b, chalkidis2017}, and an in-house dataset with sections from lease agreements.

\begin{itemize}[leftmargin=0in]
    \item[]{\bf Contract Header / Preamble} This subset contains the contract headers of the contracts of Chalkidis et al.\ \cite{chalkidis2017b, chalkidis2017}, where the goal is to identify \emph{contract titles} (3836 training/650 test element instances), \emph{parties} (6780/1250), \emph{start dates} (2210/293) and \emph{effective dates} (594/85). Start date is the date of signature and effective date is the date the agreement is enforced/activated.
    
    \item[]{\bf Applicable Law}: This subset contains the sections of the contracts of Chalkidis et al.\ \cite{chalkidis2017b, chalkidis2017} where the \emph{governing law} (2080 training/289 test) and \emph{jurisdiction} (1245/229) elements need to be identified. In case of legal disputes between parties, the governing law specifies the country or state whose laws and case law apply, while jurisdiction specifies the courts responsible to resolve any dispute.\footnote{For examples of \emph{applicable law}, see \url{https://www.lawinsider.com/clause/applicable-law}.}
    
    \item[]{\bf Lease Particulars}: This dataset contains sections from lease agreements with the following elements: address of the leased \emph{property} (2066 training/486 test), \emph{landlord} (2269/559), who owns the property, \emph{tenant} (2110/519), who rents the property, \emph{start date} (1458/346), \emph{effective date} (971/248), \emph{end date} (821/216), \emph{term} (period) of the lease (869/196), and the agreed \emph{rent amount} (1776/457).\footnote{Examples of \emph{lease particulars} can be found at \url{https://www.lawinsider.com/search?_index[0]=contract&q=lease\%20particulars}.}
\end{itemize}

\section{Experiments}
\label{experiments}

\noindent \textbf{Experimental Setup}: We used \textsc{hyperopt} and 5-fold Monte Carlo cross-validation to tune the following hyper-parameters on the training data with the following ranges: \textsc{encoder} output units \{100, 150, 200, 250, 300\}, \textsc{encoder} layers \{1, 2, 3, 4\}, batch size \{8, 12, 16, 24, 32\}, \dropout rate \{0.2, 0.3, 0.4, 0.5, 0.6\}, word \dropout rate \{0.0, 0.05, 0.1\}. 
We use the \textsc{adam} optimizer \cite{Kingma2015} with initial learning rate 1e-3. In the case of \bert, we grid-search for learning rate \{2e-5, 3e-5, 4e-5, 5e-5\}, as suggested by \citet{devlin2019}. We use the \textsc{base} versions of \bert and \legalbert
, i.e., 12 layers, 768 hidden units and 12 attention heads. All models were evaluated in terms of precision, recall, and F1-score per entity. We report mean scores on test data.

\begin{table*}[t]
\centering
\resizebox{\textwidth}{!}{
  \begin{tabular}{lccc|ccc|ccc|ccc}
  \hline
  \hline
   & \multicolumn{12}{c}{\textsc{Contract Header}} \\
  \hline
  \hline
  & \multicolumn{3}{c}{\bilstmcrf} & \multicolumn{3}{c}{\dilatedcnnscrf} & \multicolumn{3}{c}{\transformerscrf} & \multicolumn{3}{c}{\bertcrf}\\
   & P & R & F1 & P & R & F1 & P & R & F1 & P & R & F1 \\
   \cline{2-13}
   Title & \textbf{96.0} & \textbf{96.4} & \textbf{96.2} & 94.7 & 94.9 & 94.8 & 93.1 & 93.2 & 93.1 & 93.0 & 93.7 & 93.4 \\
   Party & \textbf{95.3} & \textbf{88.9} & \textbf{92.0} & 93.7 & 86.2 & 89.8 & 88.4 & 79.4 & 83.6 & 89.4 & 87.2 & 88.3\\
   S. Date & \textbf{96.8} & \textbf{97.4} & \textbf{97.1} & 91.3 & 96.6 & 93.8 & 91.3 & 92.7 & 92.0 & 94.4 & 96.3 & 95.3 \\
   E. Date & 94.6 & \textbf{96.9} & 95.7 & \textbf{96.9} & 95.1 & \textbf{95.9} & 92.0 & 88.5 & 90.1 & 86.9 & 91.3 & 89.0 \\
   \hline
   \textsc{macro-avg} & \textbf{95.7} & \textbf{94.9} & \textbf{95.2} & 94.1 & 93.2 & 93.6 & 91.2 & 88.4 & 89.7 & 90.9 & 92.1 & 91.5 \\
   \hline
  \hline
   & \multicolumn{12}{c}{\textsc{Applicable Law}} \\
  \hline
  \hline
  & \multicolumn{3}{c}{\bilstmcrf} & \multicolumn{3}{c}{\dilatedcnnscrf} & \multicolumn{3}{c}{\transformerscrf} & \multicolumn{3}{c}{\bertcrf}\\
   & P & R & F1 & P & R & F1 & P & R & F1 & P & R & F1 \\
   \cline{2-13}
   Jurisdiction & \textbf{79.7} & \textbf{72.4} & \textbf{75.9} & 69.6 & 67.6 & 68.4 & 73.6 & 58.5 & 65.0 & 74.7 & 66.8 & 70.5 \\
   Gov. Law & \textbf{98.1} & \textbf{96.3} & \textbf{97.2} & 95.1 & 92.5 & 93.8 & 98.0 & 90.3 & 94.0 & 93.8 & 92.2 & 93.0 \\
   \hline
   \textsc{macro-avg} & \textbf{88.9} & \textbf{84.4} & \textbf{86.5} & 82.3 & 80.0 & 81.1 & 85.8 & 74.4 & 79.5 & 84.3 & 79.5 & 81.8 \\
    \hline
  \hline
   & \multicolumn{12}{c}{\textsc{Lease Header}} \\
  \hline
  \hline
  & \multicolumn{3}{c}{\bilstmcrf} & \multicolumn{3}{c}{\dilatedcnnscrf} & \multicolumn{3}{c}{\transformerscrf} & \multicolumn{3}{c}{\bertcrf}\\
   & P & R & F1 & P & R & F1 & P & R & F1 & P & R & F1 \\
   \cline{2-13}
  Property & \textbf{67.0} & \textbf{65.8} & \textbf{66.2} & 61.8 & 61.8 & 61.7 & 53.9 & 50.1 & 51.8 & 54.1 & 56.1 & 55.1 \\
Landlord & \textbf{87.7} & \textbf{86.6} & \textbf{87.2} & 83.4 & 83.8 & 83.6 & 76.5 & 68.7 & 72.3 & 80.6 & 81.9 & 81.2 \\
Tenant & \textbf{90.7} & \textbf{90.9} & \textbf{90.8} & 89.7 & 87.8 & 88.7 & 81.5 & 72.4 & 76.6 & 85.1 & 87.1 & 86.1 \\
S. Date & \textbf{92.4} & \textbf{95.0} & \textbf{93.7} & 91.7 & 93.4 & 92.5 & 88.2 & 90.5 & 89.3 & 89.8 & 93.2 & 91.5 \\
E. Date & \textbf{88.7} & 90.8 & \textbf{89.7} & 81.1 & 87.5 & 84.1 & 79.9 & 71.1 & 75.2 & 85.1 &  \textbf{91.9} & 88.3 \\
T. Date & \textbf{93.9} & 85.4 & \textbf{89.3} & 91.3 & 84.2 & 87.6 & 73.2 & 67.7 & 70.2 & 86.3 & \textbf{87.0} & 86.6 \\
Period & \textbf{86.6} & \textbf{89.1} & \textbf{87.8} & 81.9 & 87.0 & 84.3 & 76.5 & 75.5 & 75.8 & 81.8 & 88.8 & 84.7 \\
Rent & \textbf{86.5} & 86.0 & \textbf{86.2} & 81.2 & 82.5 & 81.7 & 81.4 & 74.3 & 77.5 & 82.2 & \textbf{88.2} & 85.0 \\
\hline                  
\textsc{macro-avg} & \textbf{86.7} & \textbf{86.2} & \textbf{86.4} & 82.8 & 83.5 & 83.0 & 76.4 & 71.3 & 73.6 & 80.6 & 84.2 & 82.3 \\
\hline
  \end{tabular}
  }
  \caption{Results with alternative sequence encoders. \bilstm-based models are clearly better.}
  \label{tab:main}
    \vspace{-2mm}
\end{table*}

\noindent \textbf{Alternative Encoders}: Table~\ref{tab:main} reports results with different sequence encoders, always followed by a \crf layer.\footnote{With all encoders, a dense layer with a softmax activation operates on the top-level representation of each token, providing a probability distribution over the labels, which is fed to the \crf.} In these experiments except for \bertcrf, the input representation of each token is the concatenation of its word, \pos, and shape embeddings, as in our previous work \cite{chalkidis2017b}. 
Contrary to recent findings in sequence labeling \cite{strubell2017, devlin2019}, \bilstm{s} outperform \dilatedcnn{s}, stacked \transformer{s}, and \bert in all cases. Notice the particularly poor performance of \transformerscrf, which uses the same pre-trained \wordvec embeddings as \bilstmcrf and no other pre-training. 
This observation highlights the superiority of \bilstm{s} over \transformer{s}, when pre-training is limited to word embeddings, in the tasks we consider. 

More precisely, comparing \transformer{-based} methods (\transformerscrf, \bertcrf) to \bilstm{s} across  entity types, we observe that the largest performance drop in: \emph{parties} (8.4 and 3.7 F1 decrease for \transformerscrf and \bert\bertcrf, respectively), \emph{jurisdiction} (10.9, 5.4), \emph{property} (14.4, 11.1), \emph{landlord} (14.9, 6), \emph{tenant} (14.2, 4.7), and \emph{period} (12.0, 3.1). This could be attributed to the fact that, although \transformer{s} and \bert include positional embeddings and have large receptive fields, \bilstm-based models still cope better with \emph{long-term dependencies} and \emph{sequentiality}, which are important in legal documents. For example, to distinguish start and effective dates, or tenants, landlords and other parties (e.g., guarantors, etc.), or property address and other locations, one often has to consider a broader context than in generic named entity recognition. 

The particular order (sequentiality) of the context words is also important. For example, in the sentence \emph{``This Service Agreement is \underline{signed} on \textbf{\underline{February 26th, 2021}}, and \underline{effective} as of \textbf{\underline{May 1st, 2021}}.''}, the relative position of the words `signed' and `effective' is crucial to discriminate the two dates.   
\vspace{1mm}

\begin{table*}[ht!]
\centering
\footnotesize{
  \begin{tabular}{lccc|ccc|ccc}
  \hline
  \hline
   & \multicolumn{3}{c}{\textsc{Contract Header}} & \multicolumn{3}{c}{\textsc{Applicable Law}} & \multicolumn{3}{c}{\textsc{Lease Particulars}}  \\
   \hline
   \hline
   & P & R & F1 & P & R & F1 & P & R & F1 \\
   \cline{2-10}
   \bilstm{s} & 93.4 & 94.0 & 93.7 & 81.6 & 80.7 & 81.1 & 82.0 & 82.7 & 82.3 \\
   + \crf & \textbf{\underline{95.7}} & \textbf{\underline{94.9}} & \textbf{\underline{95.2}} & \textbf{\underline{88.9}} & \textbf{\underline{84.4}} & \textbf{\underline{86.5}} & \textbf{\underline{86.7}} & \textbf{\underline{86.2}} & \textbf{\underline{86.4}} \\
   \hline
   \dilatedcnn{s} & 84.2 & 88.0 & 86.0 & 68.7 & 72.7 & 70.5 
& 65.9 & 74.3 & 69.8 \\
   + \crf & \underline{94.1} & \underline{93.2} & \underline{93.6} & \underline{82.3} & \underline{80.0} & \underline{81.1} & \underline{82.8} & \underline{83.5} & \underline{83.0} \\
   \hline
   \transformer{s} & 81.8 & 86.4 & 84.0 & 54.5 & 53.9 & 54.1 & 58.0 & 64.1 & 60.8 \\
   + \crf & \underline{91.2} & \underline{88.4} & \underline{89.7} & \underline{85.8} & \underline{74.4} & \underline{79.5} & \underline{76.4} & \underline{71.3} & \underline{73.6} \\
   \hline
   \bert & 90.0 & 90.9 & 90.4 & 78.3 & 78.1 & 78.2 & 77.0 & 79.8 & 78.2 \\
   + \crf &\underline{90.9} & \underline{92.1} & \underline{91.5} & \underline{84.3} & \underline{79.5} & \underline{81.8} & \underline{80.6} & \underline{84.2} & \underline{82.3} \\ 
   \hline
  \end{tabular}
}
  \caption{Macro-averaged results with/without \crf layers. \crf{s} always improve performance.}
  \label{tab:crf}
\end{table*}

\noindent \textbf{Impact of \crf{s}}: Table~\ref{tab:crf} compares the performance of all encoders with and without \crf{s}. In each dataset, results are macro-averaged over contract element types. Similarly to prior sequence labeling studies \cite{Lample2016,strubell2017}  and unlike our previous work \cite{chalkidis2017b}, we find that \crf{s} always improve performance, especially for non-\bilstm encoders that lack recurrency (\dilatedcnn{s}, \transformer{s}).\footnote{\dilatedcnn{s} stack convolutional layers with increasingly larger strides to quickly obtain a large receptive field. \transformer{s} solely rely on additive positional embeddings and are otherwise insensitive to word order.}

\begin{table*}[ht!]
\centering
\resizebox{\textwidth}{!}{
  \begin{tabular}{lccc|ccc|ccc}
  \hline
  \hline
   & \multicolumn{3}{c}{\textsc{Contract Header}} & \multicolumn{3}{c}{\textsc{Applicable Law}} & \multicolumn{3}{c}{\textsc{Lease Particulars}}  \\
   \hline
    \hline
   & P & R & F1 & P & R & F1 & P & R & F1 \\
   \cline{2-10}
   \glove (generic) & 90.2 & 89.6 & 89.9 & 88.7 & 84.3 & 86.4 & 66.1 & 65.8 & 65.9  \\
   \wordvecword (domain-specific) & 95.7 & \textbf{95.1} & \textbf{95.4} & 89.0 & 84.1 & 86.5 & 87.0 & 86.2 & 86.6 \\
   \wordvecall (incl.\ \pos, shape) & 95.7 & 94.9 & 95.2 & 88.9 & \textbf{84.4} & 86.5 & 86.7 & 86.2 & 86.4 \\ 
   \wordvecchar & \textbf{96.1} & 94.0 & 95.0  & 89.3 & 82.2 & 85.5 & \textbf{87.8} & 86.1 & \textbf{86.9} \\
   \wordvecelmo & 95.8 & 94.8 & 95.3 & 89.3 & 84.2 & 86.7  & 86.0 & \textbf{87.5} & 86.7 \\
   \hline
  \end{tabular}
}
  \caption{Macro-averaged \bilstmcrf results with alternative input representations. Domain-specific \wordvec embeddings outperform generic \glove ones. \pos and token shape (\textsc{all}), character-based (\textsc{char}), and context-aware embeddings (\elmo) lead to no significant/consistent improvement. }
  \label{tab:features}
\end{table*}

\noindent \textbf{Alternative Feature Representations}: Table~\ref{tab:features} compares the performance of \bilstmcrf, the best encoder, with different input representations. Generic word embeddings (\glove) are vastly outperformed by domain-specific ones (\wordvecword) in two out of three datasets (\emph{contract header}, \emph{lease particulars}). Adding \pos tag and token shape embeddings (\wordvecall) does not improve overall performance (see F1 scores). Adding character-level word embeddings (\wordvecwordchar) also has no consistent or significant positive impact on F1. \elmo embeddings also do not lead to consistent noticeable improvements, possibly because the generic corpora that \elmo was trained on are very different than contracts. We suspect that in-domain knowledge is not important in \emph{applicable law}, as entities (\emph{governing law}, \emph{jurisdiction}) are mostly locations (e.g., \textsc{us} states or districts and nationality adjectives) that are properly covered in generic corpora used to pre-train \glove and \elmo.

\begin{table*}[ht!]
\centering
\footnotesize{
  \begin{tabular}{lcc|cc|cc}
  \hline
  \hline
   & \multicolumn{2}{c}{\textsc{Contract Header}} & \multicolumn{2}{c}{\textsc{Applicable Law}} & \multicolumn{2}{c}{\textsc{Lease Particulars}}  \\
   \hline
   \hline
   & Full Vocab. & Entities & Full Vocab. & Entities & Full Vocab. & Entities \\
   \cline{2-7}
     \bert & 2.10 & 1.81 & 1.71 & 1.23 & 2.55 & 2.47 \\
     \legalbert & 2.09 & 1.92 & 1.61 & 1.39 & 2.45 & 2.47 \\
     \hline
\end{tabular}
}
\caption{\emph{Word Fragmentation Ratio} (\textsc{wfr}), i.e., average ratio of sub-word units per word, for both \bert variants. We report \textsc{wfr}{s} (i) for the full vocabulary in each dataset and (ii) only for vocabulary tokens included in entities. The average \textsc{wfr} ranges from 1.2 to 2.6 and is highest in \emph{lease particulars}.}
\label{tab:word_frag}
\end{table*}

\begin{table*}[ht!]
\centering
\resizebox{\textwidth}{!}{
  \begin{tabular}{lccc|ccc|ccc}
  \hline
  \hline
   & \multicolumn{3}{c}{\textsc{Contract Header}} & \multicolumn{3}{c}{\textsc{Applicable Law}} & \multicolumn{3}{c}{\textsc{Lease Particulars}}  \\
  \hline
   & P & R & F1 & P & R & F1 & P & R & F1 \\
   \hline
    \multicolumn{10}{c}{\textsc{Stand-alone Pre-trained \transformer{s}}} \\
         \hline
    \bertcrf  & 90.9 & 92.1 & 91.5 & \underline{84.3} & \underline{79.5} & \underline{81.8} & 80.6 & \underline{84.2} & 82.3 \\
    \legalbertcrf & \underline{93.6}  & \underline{93.5} & \underline{93.5} & 81.6  & 78.8  & 80.1 & \underline{81.6} & 83.2 & \underline{82.4} \\
    \hline
    \multicolumn{10}{c}{\textsc{Feature-based Transfer Learning with \bilstmcrf (Similar to Table 3)}} \\
     \hline
       \wordvec \snowflake & \underline{95.7} & 95.1 & \underline{95.4} & 89.0 & \underline{\textbf{84.1}} & 86.5 & \underline{\textbf{87.0}} & 86.2 & \underline{86.6} \\
       \bert \snowflake & 95.1 & 92.4 & 93.7 & 90.5 & 83.9 & \textbf{\underline{87.1}} & 84.4 & 85.5 & 84.9 \\
     \legalbert \snowflake & 95.0 & \textbf{\underline{95.4}} & 95.1 & \textbf{\underline{90.6}} & 83.4 & 86.7 & 85.1 & \textbf{\underline{88.2}} & \underline{86.6} \\
    \hline
    \multicolumn{10}{c}{\textsc{Finetuning End-to-End (bert \fire, bilstm-crf \fire)}} \\
    \hline
     \bertbilstmcrf & 94.0 & 94.1 & 94.0  & \underline{88.1} & \underline{82.7} & \underline{85.2} & 83.8 & 86.4 & 85.0 \\
    \legalbertbilstmcrf & \underline{\textbf{96.3}} & \underline{95.3} & \underline{\textbf{95.8}}  & 87.3 & 82.4 & 84.7 & \underline{86.9} & \underline{87.1} & \underline{\textbf{86.9}} \\
   \hline
  \end{tabular}
}
  \caption{Macro-averaged results for \bert{-based} variants: \emph{stand-alone} (upper section) and \emph{combined} with \bilstm{s} (middle, lower). In the middle section, the \bilstm is fed with frozen \wordvec  embeddings, or frozen subword embeddings produced by \bert or \legalbert. In the lower section, \bert and \legalbert are also fine-tuned during training. Replacing \bert by \legalbert improves performance in two out of three datasets. Adding \bilstm layers further improves performance.}
  \label{tab:bert}
\end{table*}

\noindent\textbf{Why is \bert worse than \bilstm?} In order to better understand the failure of \bert in contract element extraction, we study three factors (see Table~\ref{tab:bert}): (a) the importance of in-domain language, where we compare \legalbert (a \bert model pre-trained on legal corpora \cite{chalkidis2020}) with the original pre-trained \bert; (b) the impact of recurrency, comparing with a \bilstm{-based} model that operates on top of \bert (\bertbilstmcrf) or on top of \legalbert (\legalbertbilstmcrf) in two different settings (\emph{frozen} and \emph{fine-tuned}, denoted by\snowflake and\fire); and (c) the impact of using sub-word embeddings that lead to word fragmentation (Table~\ref{tab:word_frag}), again comparing to \bertbilstmcrf and \legalbertbilstmcrf. Models relying on sub-word units need to correctly classify more tokens, contrary to models relying on words, which may lead to a performance drop. Table~\ref{tab:bert} reports results for the aforementioned models. Our observations are the following: \vspace{1mm}

\begin{itemize}[leftmargin=2.5em]
    \item Inspecting Table~\ref{tab:bert}, we observe that \legalbert leads to better performance than \bert in two out of three datasets, further highlighting the importance of in-domain knowledge in \emph{contract header} and \emph{lease particulars}, as we originally observed in Table~\ref{tab:features}. While \bert is better in the \emph{applicable law} subset, \legalbert seems to better cover companies and their roles, as suggested by the higher F1-score of \legalbertcrf comparing to \bertcrf in the corresponding entity types, i.e., \emph{party} (89.6 vs. 88.3), \emph{landlord} (85.2 vs. 81.2) and \emph{tenant} (90.5 vs. 86.1).\footnote{\emph{Per entity type} scores are not presented in Table~\ref{tab:bert} for brevity.} This observation is consistent across all methods presented in Table~\ref{tab:bert}, especially when \bilstm{s} are included. Note that \legalbert uses a vocabulary that presumably better accommodates legal language and has been pre-trained on legal corpora, while \bert uses a generic vocabulary and has been pre-trained in Wikipedia and the Children Books Corpus. 
    
    \item Inspecting the mid and lower sections of Table~\ref{tab:bert}, we observe that the methods that combine \bert or \legalbert with \bilstmcrf are comparable to \bilstmcrf relying on \wordvec embeddings, especially when models are fine-tuned end-to-end (Table~\ref{tab:bert}, lower section). These empirical results support two important findings: (a) despite their ability to capture long-term dependencies, the lack of recurrency in \transformer{-based} methods leads to worse performance, as most of the entity types greatly depend on context and sequentiality, which can be better captured by \bilstm{s}; (b) \bilstmcrf has comparable performance when operating on in-domain \wordvec or \legalbert embeddings; thus, there is no concrete evidence that the use of sub-words and the corresponding word fragmentation negatively affect performance. It seems that adding a \bilstmcrf on top of \bert or \legalbert subword embeddings alleviates any potential negative impact caused by the word fragmentation.
\end{itemize}

An important take away is that in contract element extraction, the much simpler (at least in terms of number of trainable parameters) \bilstmcrf with frozen in-domain \wordvec embeddings is very competitive to methods that employ \bert models, even when the latter are given in-domain pre-training (\legalbert) and combined with \bilstm{s} (\legalbertbilstmcrf). 

\section{Limitations and Future Work}

\bert \cite{devlin2019} was pre-trained on generic corpora, i.e, the English Wikipedia and the Children Books Corpus, while \legalbert \cite{chalkidis2020} was pre-trained on various legal corpora, comprising of legislation, regulations, court cases and contracts. Pre-training a new \bert model on larger contractual corpora,\footnote{Contractual writing is indeed very different from other genres of legal writing; for example, academic legal writing as in law journals and juridical legal writing as in court judgments \citep{Bhatia94}.} similar to those used to train \wordvec embeddings in \citet{chalkidis2017}, could possibly further improve the performance of \bert{-based} models, surpassing our currently best methods. Similarly, one can pre-train the full \textsc{bert-bilstm} encoder end-to-end, instead of plugging in randomly initialized \bilstm{s}. On the other hand, from a practical perspective, the additional computational cost needed to support the fine-tuning and inference of \bertbilstmcrf models is not justified by the minor sporadic improvements in performance. We leave this for future work, along with further investigation of the superior results of \bilstm-based methods on our datasets. The latter has possible implications in other tasks as well, given recent reports  that \transformer{-based} models under-perform in other context-sensitive (event-based) tasks \cite{kassner-schutze-2020}.

\bibliographystyle{plainnat}
\bibliography{references}

\end{document}